\documentclass[letterpaper, 10 pt, conference]{ieeeconf}  

\IEEEoverridecommandlockouts                              

\usepackage{amsmath} 

\usepackage{graphicx}
\usepackage{makecell}
\usepackage{threeparttable} 
\usepackage{subcaption}
\usepackage{xcolor}
\usepackage{booktabs}
\usepackage{multirow}

\usepackage[pagebackref = false,breaklinks,colorlinks,citecolor=green]{hyperref}

\newif\ifproofread
\proofreadfalse

\title{\LARGE \bf
An Accurate Filter-based Visual Inertial External Force Estimator via Instantaneous Accelerometer Update
}

\author{Junlin Song, Antoine Richard, and Miguel Olivares-Mendez
\thanks{This research was supported by the European Union’s Horizon 2020 project SESAME (grant agreement No 101017258). 
}
\thanks{Space Robotics (SpaceR) Research Group, Int. Centre for Security, Reliability and Trust (SnT), University of Luxembourg, Luxembourg.} 
}

\begin{document}

\maketitle
\thispagestyle{empty}
\pagestyle{empty}

\begin{abstract}

Accurate disturbance estimation is crucial for reliable robotic physical interaction.
To estimate environmental interference in a low-cost and sensorless way (without force sensor), a variety of tightly-coupled visual inertial external force estimators are proposed in the literature.
However, existing solutions may suffer from relatively low-frequency preintegration.
In this paper, a novel estimator is designed to overcome this issue via high-frequency instantaneous accelerometer update.

\end{abstract}

\section{Introduction}

\begin{figure}[htbp]
  \centering
    \begin{subfigure}[t]{0.23\textwidth}
        \centering
        \includegraphics[width=\textwidth, height=\textwidth]{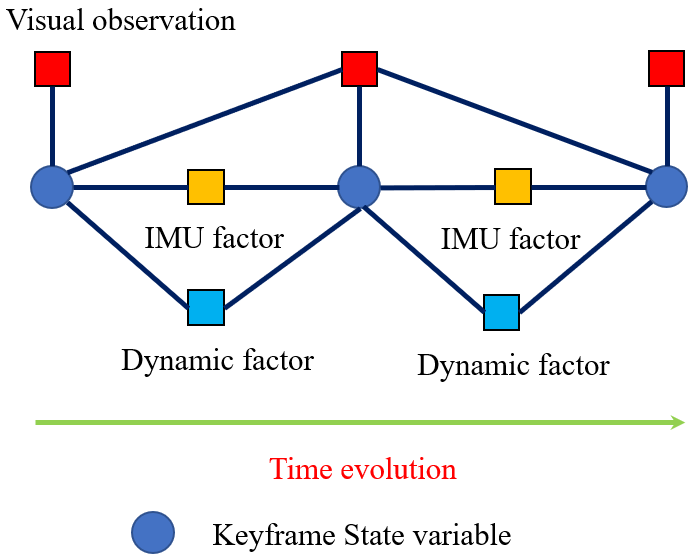}
        \caption{}
        \label{opt}
    \end{subfigure}
    \hfill
    \begin{subfigure}[t]{0.23\textwidth}
        \centering
        \includegraphics[width=\textwidth, height=\textwidth]{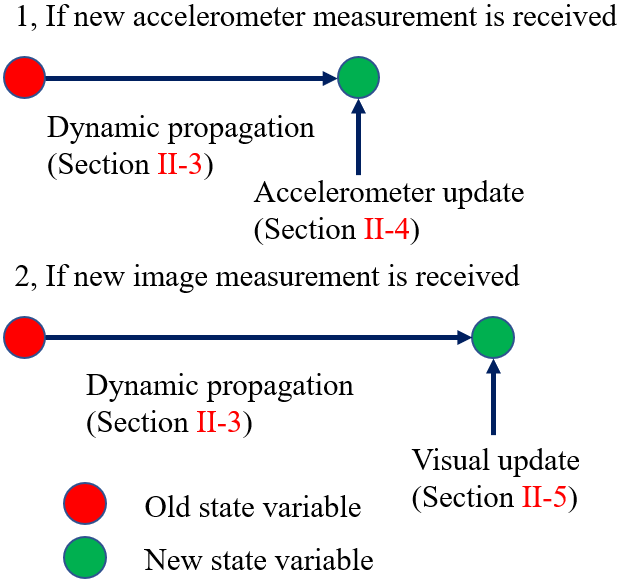}
        \caption{}
        \label{filter}
    \end{subfigure}
    \caption{(a) Factor graph framework used in \cite{nisar2019vimo, ding2021vid, cioffi2023hdvio, kang2023view}. (b) Our novel filter framework.}
\end{figure}

VIMO \cite{nisar2019vimo} is the first approach exploiting the external force within the framework of optimization-based visual inertial odometry (VIO) \cite{qin2018vins}. Dynamic measurements are preintegrated as relative motion constraints between keyframes, refering to IMU preintegration \cite{forster2016manifold}. However, the unknown external force is modeled as zero-mean Gaussian. Therefore, the estimator will be greatly affected by large or continuous external force. VID-Fusion \cite{ding2021vid} improves VIMO by exploiting the prior knowledge of external force derived from IMU and dynamic measurements. However, the external force preintegration term between adjacent keyframes is averaged to approximate the force variable. This approximation introduces estimation error. HDVIO \cite{cioffi2023hdvio} and VIEW \cite{kang2023view} employ similar factor graph framework, as shown in the Fig. \ref{opt}.

To accurately and efficiently estimate the external force, we propose a novel filter-based framework, which is presented in the Fig. \ref{filter}. Dynamic and gyroscope measurements are utilized to propagate the state variable. Accelerometer and camera measurements are used to update the state with respective frequency. The external force is updated with instantaneous accelerometer measurements without average approximation error in factor graph framework, which is prone to suffer from relatively "long-term" dynamic preintegration. We adopt a Multi-State Constraint Kalman Filter (MSCKF) \cite{mourikis2007multi} framework based on OpenVINS \cite{geneva2020openvins}. The superiority of our approach is demonstrated in the UAV case.

\section{Problem Formulation}

\subsubsection{Notation}

\begin{figure}[htbp]
  \centering
    \begin{subfigure}[t]{0.23\textwidth}
        \centering
        \includegraphics[width=\textwidth, height=0.9\textwidth]{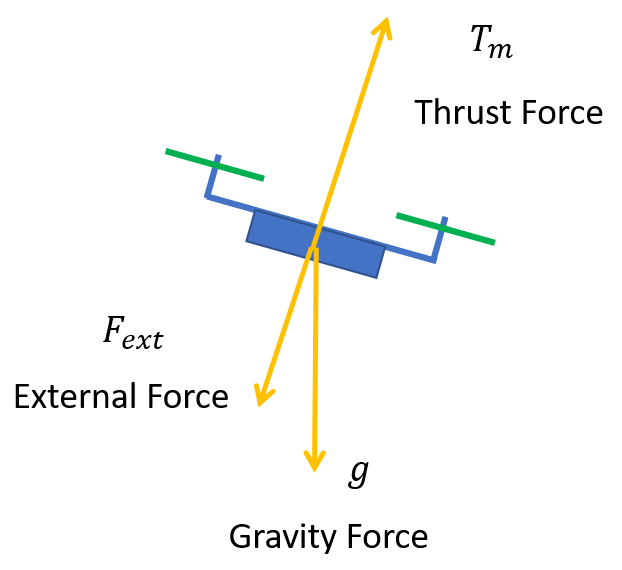}
        \caption{}
        \label{Force}
    \end{subfigure}
    \hfill
    \begin{subfigure}[t]{0.23\textwidth}
        \centering
        \includegraphics[width=\textwidth, height=0.9\textwidth]{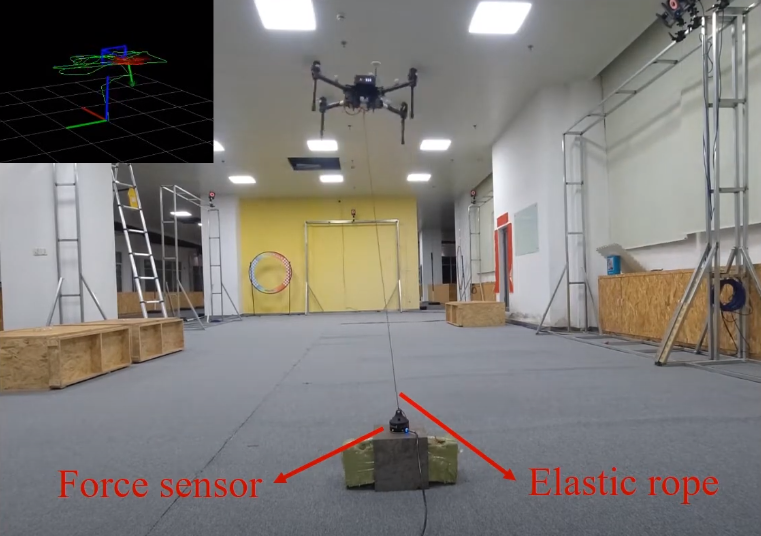}
        \caption{}
        \label{platform}
    \end{subfigure}
    \caption{(a) Mass-normalized forces of UAV. (b) Data collection platform of VID-Dataset \cite{zhang2022visual}.}
\end{figure}

There, let us define the gravity aligned world coordinate frame as $\{W\}$, the geometric central coordinate frame as $\{B\}$, the centroid frame as $\{M\}$, and the IMU frame as $\{I\}$. Frame $\{B\}$, $\{M\}$ and $\{I\}$ are assumed to be coincident, like \cite{nisar2019vimo, ding2021vid, cioffi2023hdvio}. Hereafter, only $\{I\}$ is used to refer to the body coordinate frame of UAV.

${}^W\left(  \bullet  \right)$ represents a physical quantity in frame $\{W\}$. The position of $I$ in frame $\{W\}$ is expressed as ${}^W{p_I}$. The velocity of $I$ in frame $\{W\}$ is expressed as ${}^W{v_I}$. A unit quaternion, ${}_W^Iq$, represents the attitude of frame $\{I\}$ with respect to frame $\{W\}$, and its corresponding rotation matrix is ${}_W^IR$.
The transpose of a matrix is ${\left[  \bullet  \right]^T}$. 

The force analysis of a UAV is shown in the Fig. \ref{Force}. The resultant force other than gravity and thrust is termed as external force. The mass-normalized thrust and external force are denoted as ${T_m}$ and ${F_{ext}}$ respectively. These two forces are expressed in frame $\{I\}$. $g$ is gravity vector, which is expressed in frame $\{W\}$.

\subsubsection{State Vector}

The MSCKF state vector includes the current robot state, $N$ augmented historical pose clones and $L$ augmented visual features:
\begin{equation}
    \begin{array}{l}
    x = {\left[ {\begin{array}{*{20}{c}}
    {x_I^T}&{x_c^T}&{x_f^T}
    \end{array}} \right]^T}\\
    {x_I} = {\left[ {\begin{array}{*{20}{c}}
    {{}_W^I{q^T}}&{{}^Wp_I^T}&{{}^Wv_I^T}&{b_\omega ^T}&{b_a^T}&{F_{ext}^T}
    \end{array}} \right]^T}\\
    {x_c}{\rm{ = }}{\left[ {\begin{array}{*{20}{c}}
    {x_{{c_1}}^T}& \cdots &{x_{{c_N}}^T}
    \end{array}} \right]^T}{\rm{  }}{x_{{c_i}}} = {\left[ {\begin{array}{*{20}{c}}
    {{}_W^{{I_i}}{q^T}}&{{}^Wp_{{I_i}}^T}
    \end{array}} \right]^T}\\
    {x_f}{\rm{ = }}{\left[ {\begin{array}{*{20}{c}}
    {{}^Wp_{{f_1}}^T}& \cdots &{{}^Wp_{{f_L}}^T}
    \end{array}} \right]^T}
    \end{array}
\end{equation}

Where ${x_I}$ is the current robot state, including the robot pose, velocity, the bias of IMU and external force. ${x_{{c_i}}}$ is the augmented robot pose, which is obtained by cloning the first two physical quantities of ${x_I}$ at different camera times. $N$ is known as the sliding window size, a fixed parameter. The pose clones in the sliding window are used for the triangulation of environmental visual feature points. ${}^W{p_{{f_j}}}$ is augmented feature, or SLAM feature \cite{geneva2020openvins}.

\subsubsection{Dynamic Propagation} \label{Dynamic Propagation}

The nonlinear continuous-time process model of the system can be expressed as:
\begin{equation}
    \begin{aligned}
    {}_W^I\dot q &= \frac{1}{2}\Omega \left( {{\omega _m} - {b_\omega }} \right){}_W^Iq\\
    {}^W{{\dot p}_I} &= {}^W{v_I}\\
    {}^W{{\dot v}_I} &= {}_W^I{R^T}\left( {{T_m} + {F_{ext}}} \right) + g\\
    {{\dot b}_\omega } &= {n_{{b_\omega }}},
    {{\dot b}_a} = {n_{{b_a}}},
    {{\dot F}_{ext}} = {n_F}
    \end{aligned}
\end{equation}

Where the definition of $\Omega \left( \omega  \right)$ can be found in \cite{mourikis2007multi, geneva2020openvins}. ${n_{\left[ \bullet \right]}}$  represents the zero mean Gaussian noise of $\left[  \bullet \right]$. ${x_I}$ is driven by angular velocity measurement ${\omega _m}$, and thrust measurement ${T_m}$, which is sampled as IMU frequency.

\subsubsection{Accelerometer Measurement Update}

Accelerometer measurement ${a_m}$ can observe external force.
\begin{equation}
    {a_m} = h\left( {x_I} \right) = h\left( {b_a}, F_{ext} \right) = {T_m} + {F_{ext}} + {b_a} + {n_a}
\end{equation}

Unlike factor graph framework in the Fig. \ref{opt}, the external force is updated with accelerometer frequency. We found this is an easier and more effective way than preintegration. 

\subsubsection{Visual Measurement Update}

We follow OpenVINS \cite{geneva2020openvins} for the technical details of visual update.

\section{Results}

To the best of our knowledge, in the UAV community, VID-Dataset \cite{zhang2022visual} is the first and only dataset that contains real-world visual-inertial measurement, thrust measurement, pose groundtruth and external force groundtruth. Sequence  \textit{17} and \textit{18} are chosen for validation, as only these two sequences have the groundtruth of external force. Data collection platform is shown in the Fig. \ref{platform}. The external force of UAV is dominated by the tension force of elastic rope. Therefore, the tension force measured by the force sensor can be regarded as the external force of UAV, namely ${F_{ext}}$.

\begin{figure}[htbp]
    \centering
    \includegraphics[width=0.45\textwidth]{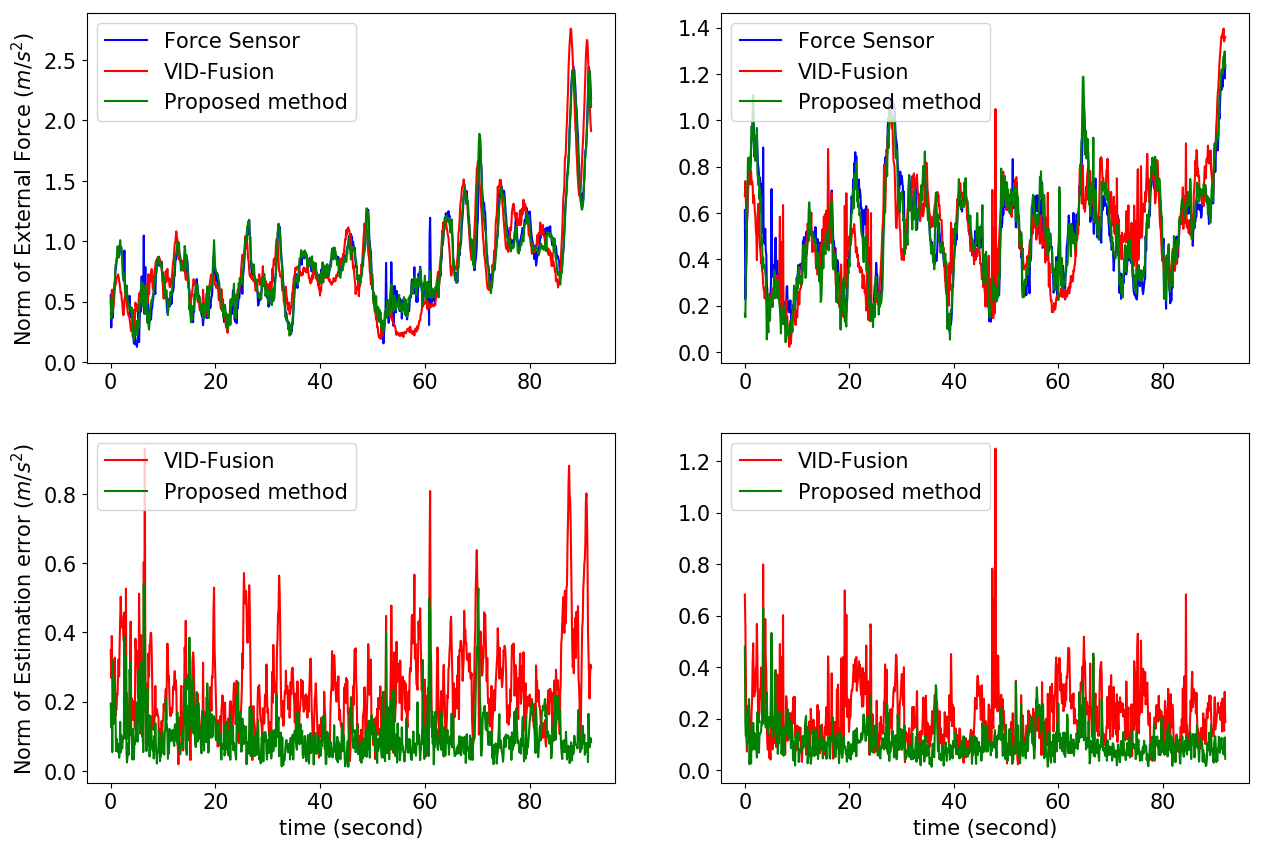}
    \caption{External force estimation: \textit{17} (left) and \textit{18} (right).}
    \label{vid_force}
\end{figure}

Estimation results of ${F_{ext}}$ are shown in the Fig. \ref{vid_force}. The quantified RMSE results are provided in the Table \ref{tab:table_force_ate}. Compared with VID-Fusion \cite{ding2021vid}, our method has significant accuracy improvement. The quantified absolute trajectory error (ATE) results are also shown in the Table \ref{tab:table_force_ate}. Our method achieves better pose estimation than VID-Fusion.

\begin{table}[]
  \caption{External force estimation RMSE ($m/{s^2}$) and ATE (m) of different algorithms (VID-Fusion, Ours) are evaluated with each sequence (seq.) of VID-Dataset.}
  \centering
  \setlength{\tabcolsep}{4pt} 
  \renewcommand{\arraystretch}{1} 
  \begin{tabular}{@{}ccccccc@{}}
    \toprule
    \multirow{2}{*}{Seq.} & \multicolumn{3}{c}{External force estimation RMSE} & \multicolumn{3}{c}{ATE} \\
    \cmidrule(lr){2-4} \cmidrule(lr){5-7}
    & {VID-Fusion} & {Ours} & {Decrease} & {VID-Fusion} & {Ours} & {Decrease} \\
    \midrule
    \textit{17}              & 0.206        & \textbf{0.072}  & 65.05\% & 0.0456        & \textbf{0.0362}  & 20.61\% \\
    \textit{18}              & 0.160        & \textbf{0.074}  & 53.75\% & 0.0788        & \textbf{0.0499}  & 36.68\% \\
    \bottomrule
  \label{tab:table_force_ate}
\end{tabular}
\end{table}

\section{Conclusion}

In this work, a novel visual inertial external force estimator is proposed to address the issue of unknown environmental interference for the robot. Compared with SOTA, the estimation accuracy is significantly improved. The methodology itself is generic and can be applied to other robotic platform than UAV, like legged robot \cite{kang2023view}, by modifying the dynamic model in Section \ref{Dynamic Propagation}. Accurate knowledge about external force would provide strong support for downstream applications, such as control and planning \cite{wu2021external, wang2022kinojgm}. 









\bibliographystyle{ieeetr}
\bibliography{bib}

\end{document}